\documentclass{vgtc}                
\ifpdf
  \pdfoutput=1\relax                   
  \usepackage{graphicx}                
  \DeclareGraphicsExtensions{.pdf,.png,.jpg,.jpeg} 
\else
  \usepackage{graphicx}                
  \DeclareGraphicsExtensions{.eps}     
\fi%

\graphicspath{{figures/}{pictures/}{images/}{./}} 

\usepackage{microtype}                 
\PassOptionsToPackage{warn}{textcomp}  
\usepackage{textcomp}                  
\usepackage{mathptmx}                  
\usepackage{times}                     
\usepackage[dvipsnames]{xcolor}
\usepackage{cite}                      
\usepackage{tabu}                      
\usepackage{booktabs}                  
\usepackage{amssymb, amsmath}
\usepackage[mathcal]{euscript}
\usepackage{mhchem}
\usepackage{isotope}
\usepackage[locale=FR]{siunitx}

\usepackage{paralist}
\usepackage[shortlabels]{enumitem}
\usepackage{verbatim}

\usepackage[font={scriptsize,sf}]{caption}
\usepackage{multicol}
\definecolor{green}{rgb}{0.0, 0.5, 0.0}
 
\definecolor{red}{rgb}{0.0, 0.0, 0.0}
\definecolor{blue}{rgb}{0.0, 0.0, 0.0}

\onlineid{8250}

\vgtccategory{Research}
\vgtcpapertype{System\vspace{-15pt}}

\vgtcinsertpkg

\title{\revise{A Machine-Learning-Aided Visual Analysis Workflow \\ for Investigating Air Pollution Data}\vspace{0pt}}

\author{
    \!\!\!\!Yun-Hsin Kuo\thanks{e-mail: \{yskuo, tfujiwara, klma\}@ucdavis.edu.}\\ %
    \scriptsize \!\!\!\!\!\!University of California, Davis\vspace{-100pt} %
\and
    \!\!\!\!\!\!\!\!\!\!\!\!\!Takanori Fujiwara\footnotemark[1]\\ %
    \scriptsize \!\!\!\!\!\!\!\!\!\!\!\!\!University of California, Davis\vspace{-100pt} %
\and 
    \!\!\!\!\!\!\!\!\!\!\!Charles C.-K. Chou\thanks{e-mail: \{ckchou@gate, cchen@stat\}.sinica.edu.tw.}\\ %
    \scriptsize \!\!\!\!\!\!\!\!\!\!\!Academia Sinica\vspace{-100pt}
\and \!\!\!\!\!\!\!\!\!Chun-houh Chen\footnotemark[2]\\
    \scriptsize \!\!\!\!\!\!\!\!\!Academia Sinica\vspace{-100pt}
\and \!\!\!\!\!\!\!\!\!\!\!\!\!Kwan-Liu Ma\footnotemark[1]\\
    \scriptsize \!\!\!\!\!\!\!\!\!\!\!\!\!University of California, Davis
    \vspace{-100pt}}

\shortauthortitle{Biv \MakeLowercase{\textit{et al.}}: Global Illumination for Fun and Profit}

\newcommand{\revise}[1]{{\color{Black}{#1}}}
\newcommand{\suggest}[1]{{\color{Black}{#1}}}

\newcommand{\MulTiDR}{MulTiDR}
\newcommand{\PM}[1]{$\mathrm{PM}_{#1}$}

\newcommand{\Scalar}[1]{\lowercase{#1}}
\newcommand{\Vector}[1]{\mathbf{\lowercase{#1}}}
\newcommand{\Matrix}[1]{\mathbf{\uppercase{#1}}}
\newcommand{\Tensor}[1]{\pmb{\mathcal{\uppercase{#1}}}}

\newcommand{\nNMFRows}{\Scalar{r}}
\newcommand{\nNMFCols}{\Scalar{c}}
\newcommand{\nFacts}{\Scalar{p}}
\newcommand{\nInsts}{\Scalar{n}}
\newcommand{\nFeats}{\Scalar{d}}
\newcommand{\nTimes}{\Scalar{t}}
\newcommand{\nInstsCLTg}{\Scalar{n}_{\mathrm{tg}}}
\newcommand{\nInstsCLBg}{\Scalar{n}_{\mathrm{bg}}}

\newcommand{\NMFOrgMat}{\Matrix{V}}
\newcommand{\NMFLatRepr}{\Matrix{W}}
\newcommand{\NMFLatFeat}{\Matrix{H}}
\newcommand{\NMFLatReprNormal}{\hat{\NMFLatRepr}}
\newcommand{\NMFLatFeatNormal}{\hat{\NMFLatFeat}}
\newcommand{\DataSymbol}{X}
\newcommand{\TensorData}{\Tensor{\DataSymbol}}
\newcommand{\UnfData}[1]{\Matrix{\DataSymbol}_{#1}}
\newcommand{\UnfDataD}{\UnfData{D}}
\newcommand{\MulTiDRDataSymbol}{U}
\newcommand{\MulTiDRTensorData}{\Tensor{\MulTiDRDataSymbol}}
\newcommand{\UnfMulTiDRData}[1]{\Matrix{\MulTiDRDataSymbol}_{#1}}
\newcommand{\UnfMulTiDRDataT}{\UnfMulTiDRData{T}}
\newcommand{\CLFeatCotrib}{\Vector{a}}

\newcommand{\ViewNameNMFResult}{pollution source}
\newcommand{\ViewNameDRScatter}{station similarity}
\newcommand{\ViewNameDRCL}{station group characteristic}
\newcommand{\ViewNameMap}{geospatial map}
\newcommand{\ViewNameTimeContrib}{pollution source contribution transition}
\newcommand{\ViewNameTrend}{air pollution transition}

\newcommand{\ViewLabelNMFResult}{a}
\newcommand{\ViewLabelDRScatter}{b}
\newcommand{\ViewLabelDRCL}{c}
\newcommand{\ViewLabelMap}{d}
\newcommand{\ViewLabelTimeContrib}{e}
\newcommand{\ViewLabelTrend}{f}

\newcommand{\norm}[1]{\left\lVert#1\right\rVert}

\abstract{
\vspace{-3pt}
Analyzing air pollution data is challenging as there are various analysis focuses from different aspects: feature (\textit{what}), space (\textit{where}), and time (\textit{when}). As in most geospatial analysis problems, besides high-dimensional features, the temporal and spatial dependencies of air pollution induce the complexity of performing analysis. Machine learning methods, such as dimensionality reduction, can extract and summarize important information of the data to lift the burden of understanding such a complicated environment. 
In this paper, we present a methodology that utilizes multiple machine learning methods to uniformly explore these aspects.
With this methodology, we develop a visual analytic system that supports a flexible analysis workflow, allowing domain experts to freely explore different aspects based on their analysis needs.
We demonstrate the capability of our system and analysis workflow supporting a variety of analysis tasks with multiple use cases.
\vspace{-2pt}
}

\keywords{Visual analytics, machine learning, analysis workflow, dimensionality reduction, matrix factorization, air pollution.}

\nocopyrightspace

\begin{document}


\firstsection{\vspace{-2pt}Introduction\vspace{-0pt}}

\maketitle

Air quality reflects the influence of natural pollution (e.g., soil dust erosion, sea breeze, or natural forest fires) and anthropogenic pollution (e.g., fuel combustion, traffic emission, or industrial activities). 
It is the indicator of the environmental health as pollution sources indirectly degrade the environment~\cite{fuzzi2015particulate}. 
Further, studies show particulate matter (PM) has associations with adverse health effects of sensitive groups~\cite{schwarze2006particulate}. 
The capability of identifying pollution contributors assists in understanding air pollution events and the decision making in environmental policies. 

In the dynamic atmospheric environment, various pollution sources, associated with different locations and time periods, contribute to the concentration of PM at one time point. Understanding air pollution requires researchers to investigate a complicated environment in the form of the spatiotemporal multivariate data, composing three major aspects: feature (\textit{what}), space (\textit{where}), and time (\textit{when})~\cite{peuquet1994s}.
Andrienko et al.~\cite{andrienko2003exploratory} accordingly derived the elementary analysis tasks with the review of visualization support. 

Researchers have developed various methods to investigate air pollution, with the focus on different aspects. 
Non-negative matrix factorization (NMF) is widely used in the field of air quality study~\cite{hopke2016review} as it extracts a variety of pollution sources from the air particle composition (e.g., consisting of ions, metals, volatile organic compounds (VOCs), and polycyclic aromatic hydrocarbons (PAHs)). 
Anomaly detection is being employed to identify potential air pollution events from the time aspect~\cite{chen2017adf, chuang2017simulation}.
Supervised models, such as CMAQ~\cite{byun2006review}, simulate atmospheric diffusion and chemical reactions to capture the environmental complexity. Juxtaposing geospatial maps often delivers the reproduced spatial distribution to let analysts retrospectively or prospectively study air pollution~\cite{chuang2017simulation}. 
These methods help conduct in-depth analysis; however, they only provide insights from two of the three aspects at most.

%
With the assistance of interactive visualizations, we can effectively connect different aspects of the data. 
Several visualizations are designed to uncover patterns from the relationships among pollutants~\cite{qu2007visual, ren2020visual}, while some computational analysis methods are good to aid visual exploration.
Interactive machine learning (ML) methods (e.g., clustering methods) have been used to visually compare different regions~\cite{engel2012visual, zhou2017visual, zhang2019airinsight}. 
However, these existing approaches still do not provide effective visual analytics support to uniformly explore the data from the aforementioned three aspects.


In this paper, we present a methodology that effectively links the three aspects (i.e., feature/variable, space, and time) of \revise{outdoor} air pollution data for interactive analysis. To achieve this, we introduce an ML pipeline that aims to provide a unified view of the data as well as to support flexible exploration over different combinations of the three aspects. 
Within this pipeline, we employ NMF to extract the air pollution sources, incorporate correlation analysis to assist the source interpretation, and use the factorization result to further explore the time and space aspects.
To highlight each region's uniqueness in terms of the association with the air pollution sources, we further adopt contrastive learning, which is an emerging ML scheme designed to extract salient/unique patterns in one dataset relative to the other~\cite{abid2018exploring}. 
We develop a visual analytic system supporting a flexible analysis workflow to allow free exploration and top-down analysis. 
By linking the three aspects, we are able to uncover the hidden air pollution events that happened at a specific location and time and identify the contributing pollution sources.

Our main contributions include: 
(1) introducing an ML pipeline that exploits existing dimensionality reduction (DR) and clustering methods to investigate the three aspects of the data; 
(2) prototyping a visual analytic system that coordinates linked visualizations to provide a flexible analysis workflow; 
(3) demonstrating the flexible analysis workflow that aids a variety of analysis tasks with multiple use cases using real-world air pollution datasets.

\vspace{-2pt}
\section{Background and Related Work}
\label{sec:relatedwork}

Our work aids data-driven air pollution analysis with interactive visualizations.
We first describe typical air pollution data and then discuss representative research on the related topics.


\subsection{Air Pollution Data}
\label{sec:data}


Air pollution is usually measured with six common air pollutants (called criteria air pollutants~\cite{criteria_air}):\,\PM{2.5},\,\PM{10},\,\ce{NO2}, \ce{SO2}, \ce{O3}, and \ce{CO}. 
Among these, PM is a mixture of heterogeneous small particles from multiple \textit{pollution sources}.
Along with the criteria air pollutants, researchers measure a variety of chemical species, including ions, metals, VOCs, and PAHs, to further understand the details of PM and identify the sources of PM.
For example, if we identify the pollution source highly relevant to \ce{Na+} and \ce{Cl-}, the air pollution is likely caused by strong sea breeze\suggest{s as they deliver} these ions composing salt (i.e., \ce{NaCl}). 
As we discuss in \autoref{sec:data_driven_analysis}, for this identification process, researchers often use a matrix factorization method, such as factor analysis, principal component analysis (PCA), or NMF. 
Note that the cost of using various sensors to measure many different chemical species is often more expensive than using sensors of the criteria air pollutants~\cite{measuretech}; thus, the chemical species tend to be only measured at limited locations. 



\noindent\textbf{Concrete example of air pollution data.}
In the ensuing sections, we describe our visual analytics system while analyzing air pollution data obtained at 12 monitoring sites in the central region of Taiwan. 
Each monitoring site has one meteorological station and one air quality station. 
The meteorological station samples wind speed, wind direction, temperature, and relative humidity on an hourly basis. 
The air quality station collects two different types of data at different sampling rates. 
The first type includes values of the criteria air pollutants plus \ce{NO} recorded hourly. 
The second type consists of the concentration values of 49 chemical species, which include 8 ions (e.g., \ce{Na+} and \ce{Cl-}), 
36 metals (e.g., \ce{Al} and \ce{Fe}), organic carbon, elemental carbon, Levoglucosan, Mannosan, and Galactosan, \textcolor{blue}{all of which are the mean value for 12 hours and recorded every 8 am and 8 pm.}
In addition, we have GPS coordinates of the monitoring sites. 
To supplement our analysis, we also incorporate a dataset from a large amount of \PM{2.5} sensors in the same region. 
The dataset consists of values of \PM{2.5} recorded at 537 AirBox sensors~\cite{airbox} hourly, along with the sensors' GPS coordinates.
All of the data were collected from March 12th to 31st, 2018.  

\vspace{-2pt}
\subsection{Data-Driven Air Pollution Analysis}
\label{sec:data_driven_analysis}
\vspace{-2pt}


For air pollution data analysis, NMF (also known as positive matrix factorization) is one of the most widely used ML methods~\cite{hopke2016review}. 
NMF decomposes a matrix of air pollution data (where rows and columns represent samples and chemical species, respectively) into two smaller matrices. 
By referring to these two matrices, analysts can identify sources/factors of air pollution~\cite{brown2007source} (see \autoref{sec:source_extraction} for the details).
For example, Hasheminassab et al.~\cite{hasheminassab2014long} studied the impact \suggest{of} the vehicular emissions \suggest{on} air pollution sources.
Gon et al.~\cite{gon2018source} compared pollution source compositions at different geospatial regions.
Liao et al.~\cite{liao2020vertical} further compared the compositions at different levels of altitudes. 


As seen \suggest{with} the emergence of urban computing~\cite{zheng2014urban}, low-cost particle sensing devices increase the granularity of air pollution data, enabling advanced data-driven analysis, such as \revise{ML on air quality prediction~\cite{zheng2013u, yi2018deep} and 
pollution event classification~\cite{saad2015classifying, fang2016airsense}.}
However, as low-cost sensing devices \suggest{may} easily cause errors and malfunctions, researchers developed monitoring methods to ensure \suggest{data quality }~\cite{chen2017adf}.
As described in the ensuing subsection, visualization also can help review and analyze large air pollution data while considering potential problems in the quality of the data.

We use ML methods commonly used in existing works (e.g., NMF) and enhance analysis flexibility and interpretability of ML results by coupling contrastive learning and interactive visualization.

\vspace{-2pt}
\subsection{Visualizations for Air Pollution Analysis}
\vspace{-2pt}

Air pollution data is often multivariate \revise{and} spatiotemporal. 
There is a large collection of existing works on multivariate, spatiotemporal visualizations, as referred by recent literature surveys~\cite{zheng2016visual,bach2017descriptive}.
We focus on works specifically designed for outdoor air pollution analysis (e.g., the air pollution source identification).

\suggest{Analysis of air pollution data often requires visual exploration from each of the three aspects, feature, space, and time, and their combinations.}
For example, Qu et al.~\cite{qu2007visual} enhanced parallel coordinates and polar-coordinate-based scatterplots to inform relationships among air quality relevant features.
Ren et al.~\cite{ren2020visual} summarized the air pollutant propagation by constructing a network for each time point based on the similarities of air pollutant distributions per location.
They further visualized the distribution changes for each network community (i.e., group of cities) over time. 
These works and others~\cite{lu2017interactive,li2016visualization} highly relied on visualizations to find patterns rather than utilizing ML methods. 

Several works aided visual exploration with ML methods.
Engel et al.~\cite{engel2012visual} enhanced NMF by adding regularization terms that can be used to reduce the influence \suggest{of noise and uncertainty} in the data. 
They also provided a visual interface to adjust NMF results while \suggest{keeping} the expert in the loop. 
Zhou et al.~\cite{zhou2017visual} utilized DR and clustering methods 
to group locations \suggest{with} similar air pollutant distributions. 
Zhang et~al.~\cite{zhang2019airinsight} took a similar approach to group samples.
Deng et al.~\cite{deng2019airvis} constructed directed networks that represent air pollution propagation between different locations and employed frequent subgraph mining to help review the propagation patterns.
Others employed temporal data mining methods, such as  time-series analysis methods~\cite{qu2020airexplorer,guo2019visual} and recurrent neural networks~\cite{liu2021aqeyes}.

Our work analyzes air pollution data from the three aspects while \suggest{combining} interactive visualization and ML.
We introduce an analysis pipeline that enables an integrated analysis of the three aspects, as described in \autoref{sec:methodology}.
Also, while the existing visual analytics systems only support predetermined (one-directional) analysis workflow, our system supports a more flexible analysis workflow to explore each of the three aspects or the combinations of them.

\vspace{-2pt}
\section{Design Goals}
\vspace{-1pt}
\label{sec:requirements}

As we discussed through our literature survey, analysis of air pollution data usually requires the exploration from the three aspects: feature, space, and time.  
Our general goal is to support flexible analysis from each individual aspect as well as each of the combinations of them (e.g., feature and space).
Also, we aim to effectively help domain experts perform their common tasks (e.g., pollution source identification).
The detailed design goals (\textbf{DGs}) of our visual analytics system are listed below.
\revise{While we derive \textbf{DGs} from the literature survey, they are also validated by the domain experts during the expert interview, as described in \autoref{sec:expertreview}.}



\textbf{DG1: Supporting pollution source identification.}
The identification of pollution sources is the most common task related to the aspect of features\revise{~\cite{hopke2016review}}.
The system should support a matrix factorization method that domain experts are familiar with (specifically, NMF). 
Also, we should provide visualizations that help them interpret the matrix factorization results with the contexts of measured air pollutants (e.g., \PM{2.5}) and chemical species (e.g., \ce{Na+}).
 
\textbf{DG2: Aiding pollution event identification and explanation.}
An important analysis task related to the time aspect is identifying when the cause of air pollution occurred---we call this cause \textit{pollution event}\revise{~\cite{chuang2017simulation}}.
The system should aid domain experts in performing this task as well as understanding pollution events with auxiliary information, such as related locations and lasting periods. 

\textbf{DG3: Characterizing geospatial regions.}
Similar pollution patterns are often observed in adjacent stations. 
As seen in existing works\revise{~\cite{gon2018source}}, based on the patterns, domain experts often want to form regions encompassing multiple stations and understand their characteristics (e.g., region A is influenced by sea breeze\suggest{s}). 
The system should support this grouping and characterization process.

\textbf{DG4: Enabling effective exploration in pairs of the aspects.}
To thoroughly analyze air pollution data, exploration from the combinations of the three aspects is necessary\revise{~\cite{gon2018source, hasheminassab2014long, chuang2017simulation}}. 
However, simultaneously examining multiple aspects can be a challenging task as it involves excessive information.
Our system should provide computational analysis and visualization support\suggest{, allowing} users to effectively explore the complex information. 

\textbf{DG5: Providing analysis flexibility.}
As air pollution data analysis often involves various analysis focuses from different aspects\revise{~\cite{gao2018temporal}}, the system should be able to assist a wide variety of analysis needs, which could 
\suggest{differ between} each domain expert.
\suggest{Furthermore, a domain expert often changes their needs during their analysis.}
Thus, the system should support multiple different analysis workflows with a flexible visual interface. 
The experts should be able to focus on a particular analysis, and seamlessly move on to a different analysis based on their intermediate findings.

\begin{figure*}[tb]
	\centering
    \includegraphics[width=\linewidth,height=0.42\linewidth]{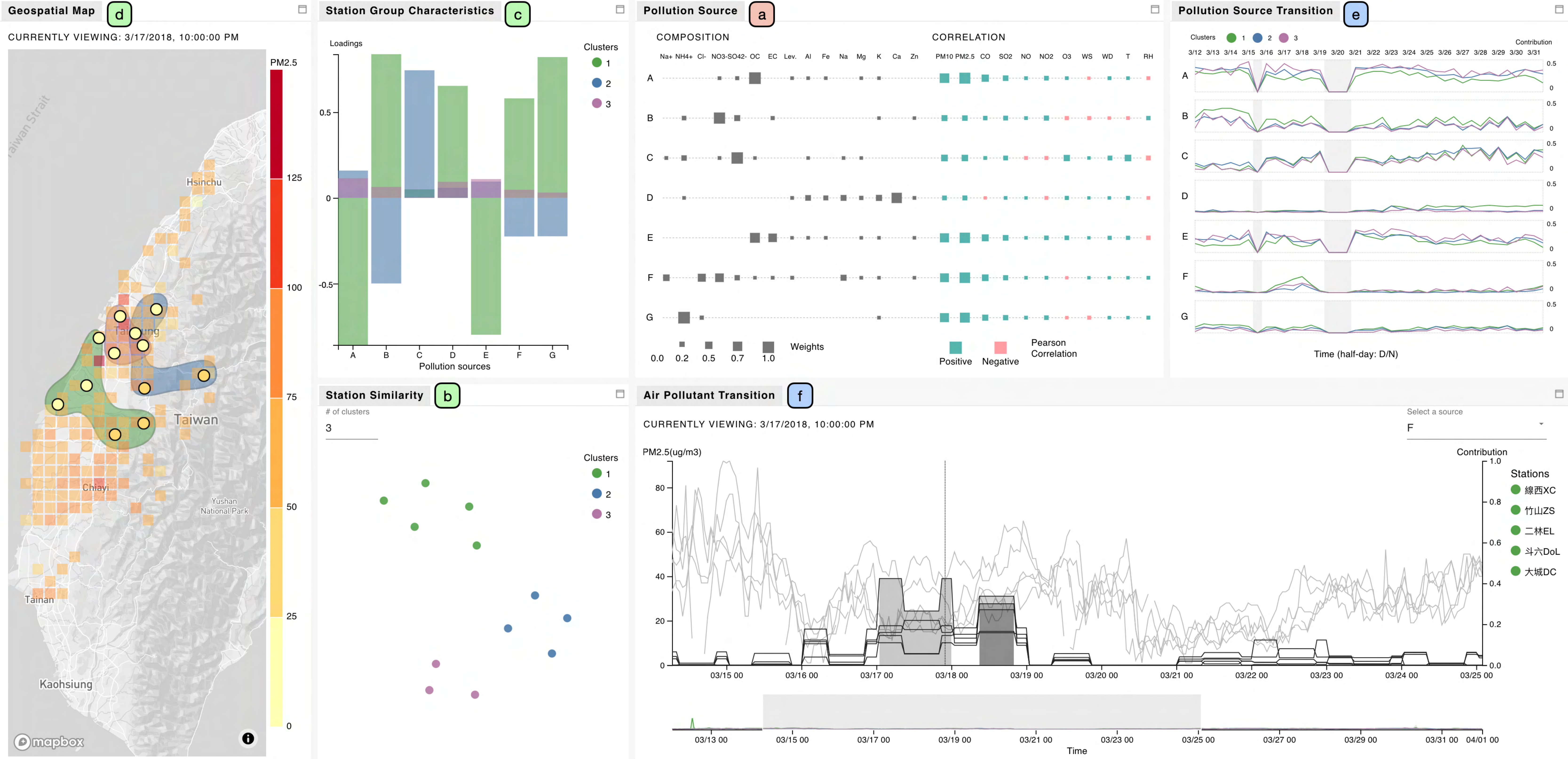}
    \caption{
    A screenshot of our system's visual interface. 
    Here we analyze the air pollution collected at the central region of Taiwan. 
    (\ViewLabelNMFResult) the {\ViewNameNMFResult} view depicts the composition of pollution source identified by NMF (left) as well as the correlations with air pollutants and meteorological measures (right). (\ViewLabelDRScatter) the {\ViewNameDRScatter} view shows the similarity of stations and their cluster ID, which are extracted with the \MulTiDR{} framework. (\ViewLabelDRCL) the {\ViewNameDRCL} view visualizes the influence of each pollution source on each station group/cluster. (\ViewLabelMap) the {\ViewNameMap} view conveys geological locations of stations with their cluster ID. (\ViewLabelTimeContrib) the {\ViewNameTimeContrib} view informs transitions of pollution source contributions. (\ViewLabelTrend) the {\ViewNameTrend} view displays selected stations' PM$_{2.5}$ values with a pollution source contribution.
    }
	\label{fig:overview}
\end{figure*}

\begin{figure*}
 \centering 
 \includegraphics[width=\linewidth,height=0.355\linewidth]{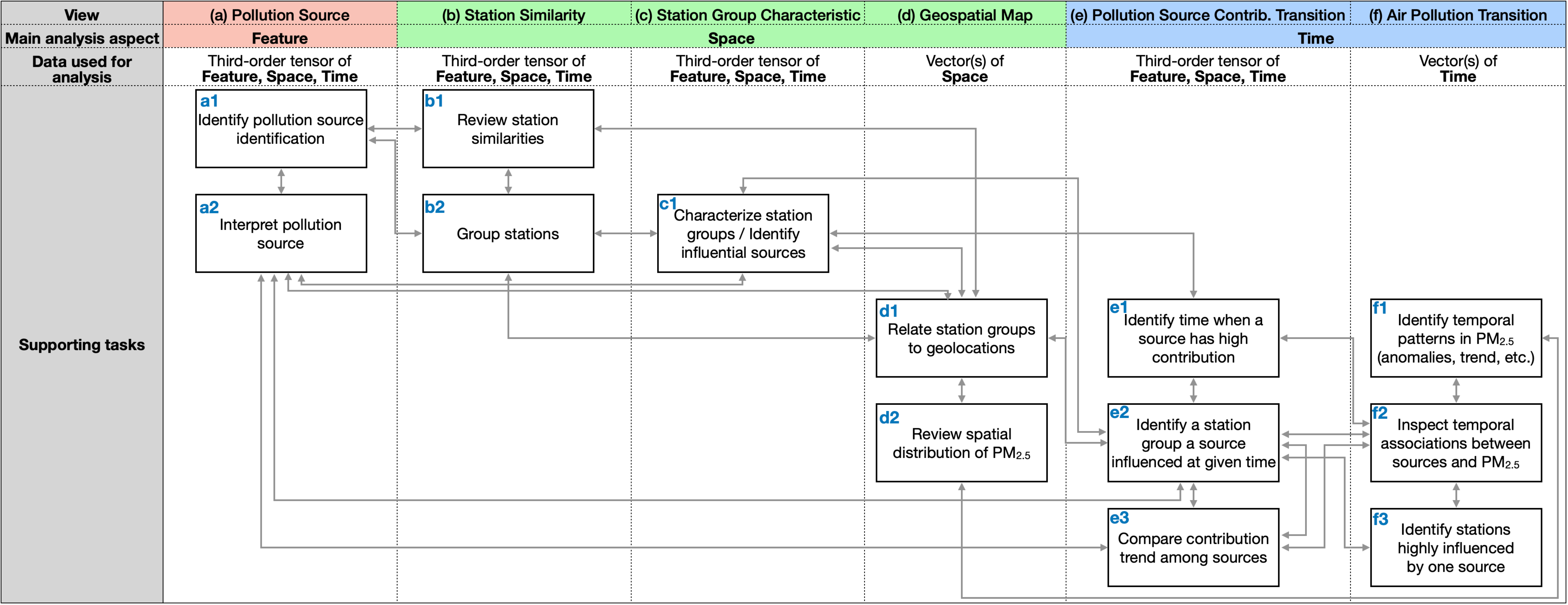}
 \caption{
 The flexible analysis workflow. 
 Multiple views collectively support the analysis from the feature, space, and time aspects. 
 Each view has a main analysis target (e.g., the {\ViewNameNMFResult} view is mainly for analyses from the feature aspect); however, most views consider all the aspects when producing visualizations (e.g., the {\ViewNameNMFResult} view uses all the three aspects), enabling seamless analyses from multiple aspects.
 Each view supports one or multiple tasks (e.g., the {\ViewNameNMFResult} view supports \texttt{Tasks a1} and \texttt{a2}). 
 Each supported task can be performed after the other and is highly liked with many other tasks; consequently, the flexible analysis workflow encompasses various subworkflows (e.g.,  performing \texttt{Tasks a1}, \texttt{a2}, and then \texttt{d1}). While here covers the main task dependencies, each task could have more links to others. 
 }
 \vspace{-1pt}
 \label{fig:workflow}
\end{figure*}

\section{Methodology}
\label{sec:methodology}

Based on the \textbf{DGs}, we design a visual analytics system (\autoref{fig:overview}) that supports a flexible analysis workflow (\autoref{fig:workflow}) to analyze complex air pollution data from multiple aspects while utilizing ML methods. 
As shown in \autoref{fig:overview}, the system consists of six views: 
(\ViewLabelNMFResult) {\ViewNameNMFResult}, (\ViewLabelDRScatter) {\ViewNameDRScatter}, (\ViewLabelDRCL) {\ViewNameDRCL}, (\ViewLabelMap) {\ViewNameMap}, (\ViewLabelTimeContrib) {\ViewNameTimeContrib}, and (\ViewLabelTrend) {\ViewNameTrend}  views.
Each view is designed to support the exploration from a different aspect. 
For example, while the {\ViewNameNMFResult} view is mainly \revise{used} to review air pollution sources (i.e., the feature aspect), the {\ViewNameDRCL} view can be used to identify pollution sources that have high influence on each region (i.e., the space aspect while considering the features). 
The supported explorations (or analysis tasks) are listed in the analysis workflow (\autoref{fig:workflow}).
With the fully linked coordination of the views, the system \revise{is able} to support the flexible analysis workflow.
We provide a supplementary video to demonstrate the system's functionalities.

\vspace{-1pt}
\subsection{Flexible Analysis Workflow}
\vspace{-1pt}
\label{sec:flexible_workflow}

As seen in \autoref{fig:workflow}, the flexible analysis workflow encompasses multiple different subworkflows (\textbf{DG5}). 
For example, if domain experts want to understand geographical regions' characteristics, they can first (i) compare the {\ViewNameDRScatter} and {\ViewNameMap} views to know how stations are grouped together (\texttt{Tasks\,b2},\,\texttt{d1}), (ii) identify influential pollution sources from the {\ViewNameDRCL} view (\texttt{Task\,c1}), and then (iii) interpret the pollution sources with the chemical species by looking at the {\ViewNameNMFResult} view (\texttt{Task\,a2}). 

Another \suggest{subworkflow} is to comprehensively understand pollution sources from multiple aspects\suggest{. First, we} (i) identify and interpreting pollution sources (\texttt{Tasks a1}, \texttt{a2}) and then (ii) associate the pollution sources and regions with the tasks stated in the first subworkflow example (i.e., \texttt{Tasks b2}, \texttt{c1}, and \texttt{d1}).
Afterwards, we can (iii) assess the changes of the influence of each pollution source over time with the {\ViewNameTimeContrib} view (\texttt{Task e1}) and (iv) see the relationships with \PM{2.5} (\texttt{Task f2}).

We have explained only two subworkflows, but there are myriads of subworkflows domain experts can follow. 
As an analysis target constantly changes while moving back and forward to each of the tasks, we design the system interface to be flexible for rearranging the layout. 
The system allows domain experts to easily switch the positions of views and focus on a particular view by changing its window size.
In the ensuing subsections, we use the second subworkflow explained above in order to introduce each view and analysis component with a concrete analysis example.
We demonstrate more analysis examples with different subworkflows in \autoref{sec:cases}.





\vspace{-1pt}
\subsection{Pollution Source Identification and Interpretation}
\label{sec:source_extraction}

One of our core contributions is an ML pipeline (\autoref{fig:pipeline}) designed to summarize air pollution data from all the three aspects (\textbf{DG4}). 
The ML pipeline consists of two parts: (a) pollution source identification and interpretation using NMF (\textbf{DG1}, \textbf{DG4}) and (b) region summary and characterization using the \MulTiDR{} framework~\cite{fujiwara2020visual} (\textbf{DG3}, \textbf{DG4}).
Here we describe components related to the first part: NMF and visualizations shown in the {\ViewNameNMFResult} view (\autoref{fig:overview}-{\ViewLabelNMFResult}1, {\ViewLabelNMFResult}2), which help interpret the NMF result. 

\vspace{3pt}
\noindent
\textbf{Non-negative matrix factorization (NMF).}
As its name indicates, NMF decomposes a matrix into two matrices where all elements have non-negative numbers. 
More formally, given a matrix $\smash{\NMFOrgMat \in \mathbb{R}_{+}^{\nNMFRows \times \nNMFCols}}$ ($\nNMFRows$ and $\nNMFCols$ are the numbers of rows/observations and columns/dimensions) and the number of factors $\nFacts$, NMF produces two non-negative matrices $\smash{\NMFLatRepr \in \mathbb{R}_{+}^{\nNMFRows \times \nFacts}}$ and $\smash{\NMFLatFeat \in \mathbb{R}_{+}^{\nFacts \times \nNMFCols}}$ such that their product, $\NMFLatRepr \NMFLatFeat$, approximates $\NMFOrgMat$ by minimizing $\smash{\norm{ \NMFOrgMat - \NMFLatRepr \NMFLatFeat }^2_F}$ (note $\smash{\norm{ \cdot }_F}$ is the Frobenius norm). 
Similar to those generated by PCA, $\NMFLatRepr$ and $\NMFLatFeat$ are latent/lower-dimensional representations of observations and latent features/factors (corresponding to components in PCA), respectively.
The strength of NMF over methods that may produce negative numbers in $\NMFLatFeat$ (e.g., PCA) is in its interpretability of $\NMFLatFeat$.
By restricting loadings (i.e., elements in each row of $\NMFLatFeat$) to non-negative values, each latent feature can be simply interpreted as a weighted accumulation of original features that have non-zero loadings.


In our case, we use NMF to identify pollution sources from the data of chemical species related to air pollution (\textbf{DG1}), which contains the information on the three aspects.
This data can be represented as a third-order tensor  $\smash{\TensorData \in \mathbb{R}^{\nTimes \times \nInsts \times \nFeats}}$ where $\nTimes$, $\nInsts$, and $\nFeats$ are the numbers of timestamps, stations, and features, respectively.
For instance, when using the data obtained from the 12 sites in Taiwan (described in \autoref{sec:data}), $\nTimes \!\!=\!\! 40$, $\nInsts \!\!=\!\! 12$, and $\nFeats \!\!=\!\! 49$.
As NMF can be only performed on a matrix, similar to the work by Fujiwara et al.~\cite{fujiwara2020visual}, we first apply tensor unfolding to $\TensorData$ (\autoref{fig:pipeline}-a1).
Tensor unfolding reshapes $\TensorData$ to a matrix $\smash{\UnfDataD}$ of $\nTimes \nInsts$ rows and $\nFeats$ columns (i.e., $\smash{\UnfDataD \in \mathbb{R}_{+}^{\nTimes \nInsts \times \nFeats}}$) by arranging all vectors of $\nFeats$ length obtained through the slicing of $\TensorData$ along both timestamps and stations.
We then apply NMF to $\smash{\UnfDataD}$ (\autoref{fig:pipeline}-a2).
Unlike PCA, all latent features in the NMF results are highly influenced by the selection of $\nFacts$. 
We follow a common approach taken by domain experts to select $\nFacts$: manually searching a small number $\nFacts$ so that each latent feature likely shows a distinct pollution source~\revise{\cite{liao2020vertical}} (e.g., $\nFacts \!\!=\!\! 7$ in \autoref{fig:overview}-{\ViewLabelNMFResult}1). 

$\smash{\NMFLatFeat \in \mathbb{R}_{+}^{\nFacts \times \nFeats}}$ derived from NMF on $\smash{\UnfDataD}$ contains numerical mappings between original features (i.e., chemical species) and latent features (i.e., pollution sources).
For example, when $\NMFLatFeat$'s columns correspond to $\smash{\ce{Na+}}$, $\smash{\ce{Cl-}}$, $\smash{\mathrm{NH}_4^+}$, and $\smash{\mathrm{NO}_3^-}$ and one row vector of $\NMFLatFeat$ is $\smash{[0.5, 0.5, 0, 0]}$, this row (i.e., pollution source) is likely related to salt (\ce{NaCl}). 
On the other hand, $\smash{\NMFLatRepr \in \mathbb{R}_{+}^{\nTimes \nInsts \times \nFacts}}$ shows numerical mappings between each pollution source and each pair of $\nInsts$ stations and $\nTimes$ timestamps.
By looking at these mappings, we can identify pollution sources that highly contribute to a certain station-timestamp pair. 

As we mainly want to know which chemical species highly associate with each pollution source, we further apply row-wise $l2$-normalization to $\NMFLatFeat$ (\autoref{fig:pipeline}-a3), which emphasizes high values while de-emphasizing low values.
Similarly, we apply row-wise $l1$-normalization to $\NMFLatRepr$ in order to preprocess this matrix for the ensuing process in the ML pipeline and to compare how the pollution sources contribute to different sensor-timestamp pairs (\autoref{fig:pipeline}-a4). 
We denote the normalized $\smash{\NMFLatFeat}$ and $\smash{\NMFLatRepr}$ as $\NMFLatFeatNormal$ and $\NMFLatReprNormal$, respectively.
In the rest of this paper, following existing works on air pollution~\cite{hopke2016review}, we call each element value in $\NMFLatFeatNormal$ \textit{concentration} of a chemical species in a pollution source; each element value in $\smash{\NMFLatReprNormal}$ \textit{contribution} from a pollution source to a station, respectively. 
$\smash{\NMFLatReprNormal}$ is also used for the rest of the analysis pipeline, as described in \autoref{sec:dr}. 

\vspace{3pt}
\noindent
\textbf{Visualizations to help interpret NMF results.}
The {\ViewNameNMFResult} view provides two visualizations to aid the interpretation of the pollution sources identified by NMF. 
While the first visualization (\autoref{fig:overview}-{\ViewLabelNMFResult}(left)) depicts the information of $\smash{\NMFLatFeatNormal}$, while the second visualization (\autoref{fig:overview}-{\ViewLabelNMFResult}(right)) presents the correlations between the pollution sources and air pollutants (e.g., \PM{2.5}) as well as meteorological measures (e.g., temperature).

In \autoref{fig:overview}-{\ViewLabelNMFResult}(left), each row and column corresponds to a pollution source and a chemical species, respectively. 
We label each pollution source with a capitalized letter starting from \texttt{A} (e.g., seven pollution sources \texttt{A}--\texttt{G}). 
Because the number of chemical species can be large (e.g., 49 in our data), by default, we show the top-15 chemical species ranked by the column-wise total of $\NMFLatFeatNormal$. 
We visually encode each element of $\NMFLatFeatNormal$ with a different size of a square.
We select this encoding so as to emphasize large values rather than small ones (the same motivation as $l2$-normalization applied on $\NMFLatFeat$). 
\suggest{The} actual value also can be checked by hovering over each square.



For \autoref{fig:overview}-{\ViewLabelNMFResult}(right), we employ a similar design to \autoref{fig:overview}-{\ViewLabelNMFResult}(left).
Instead of a chemical species, each column corresponds to an air pollutant or a meteorological measure.
Also, we use different colors to differentiate positive and negative correlations (positive: teal, negative: pink). 
By default, a correlation value shows the Pearson's correlation coefficient between two sets of values of the corresponding pair of pollution source and air pollutant/meteorological measure recorded at $\nInsts$ stations and $\nTimes$ timestamps (i.e., each set has $\nInsts \nTimes$ values).


From these visualizations, for example, in \autoref{fig:overview}-{\ViewLabelNMFResult}(left), we can observe that \ce{Na+} and \ce{Cl-} ions have a high concentration only in \texttt{Source F}. 
This source could be sea breezes as the two ions are commonly found in the composition of salt.
We can further confirm this by reviewing other views (e.g., checking whether or not \texttt{Source F} highly contributes to a region close to the ocean) while incorporating domain knowledge.
From \autoref{fig:overview}-{\ViewLabelNMFResult}(right), we can see \texttt{Source D} has the small-size squares for all air pollutants.
Thus, when compared with the other sources, \texttt{Source D} likely had less influence on the air pollution during the studied period.
We have identified the air pollution sources and interpreted some of them (\texttt{Tasks a1}, \texttt{a2}). 
Using other views introduced in the following sections, we further analyze \texttt{Source F}, which is likely related to sea breeze.

\vspace{-1pt}
\subsection{Region Summarization and Characterization}
\vspace{-1pt}
\label{sec:region}

Our ML pipeline helps to construct groups of stations based on temporal similarities in their pollution sources to make spatial related analyses easier (\textbf{DG3}, \textbf{DG4}).
Even though our data has a relatively small number of stations (12 stations), it is not trivial to compare contributions of pollution sources to these stations all at once (e.g., with many bar charts). 
This analysis scalability issue becomes even worse if we have more stations, as described in \autoref{sec:case-us}.
Using contrastive learning~\cite{fujiwara2020visual}, the ML pipeline also aids the characterization of the groups.
We perform the grouping and characterization based on the identified air pollution sources since research on air pollution often targets on analyses related to the pollution sources.
\suggest{The ML pipeline addresses this functionality by integrating the NMF process into the \MulTiDR{} framework~\cite{fujiwara2020visual}.}

\vspace{-2pt}
\subsubsection{Region Summarization}
\vspace{-1pt}
\label{sec:dr}

\begin{figure}[tb]
 \centering 
 \includegraphics[width=\linewidth]{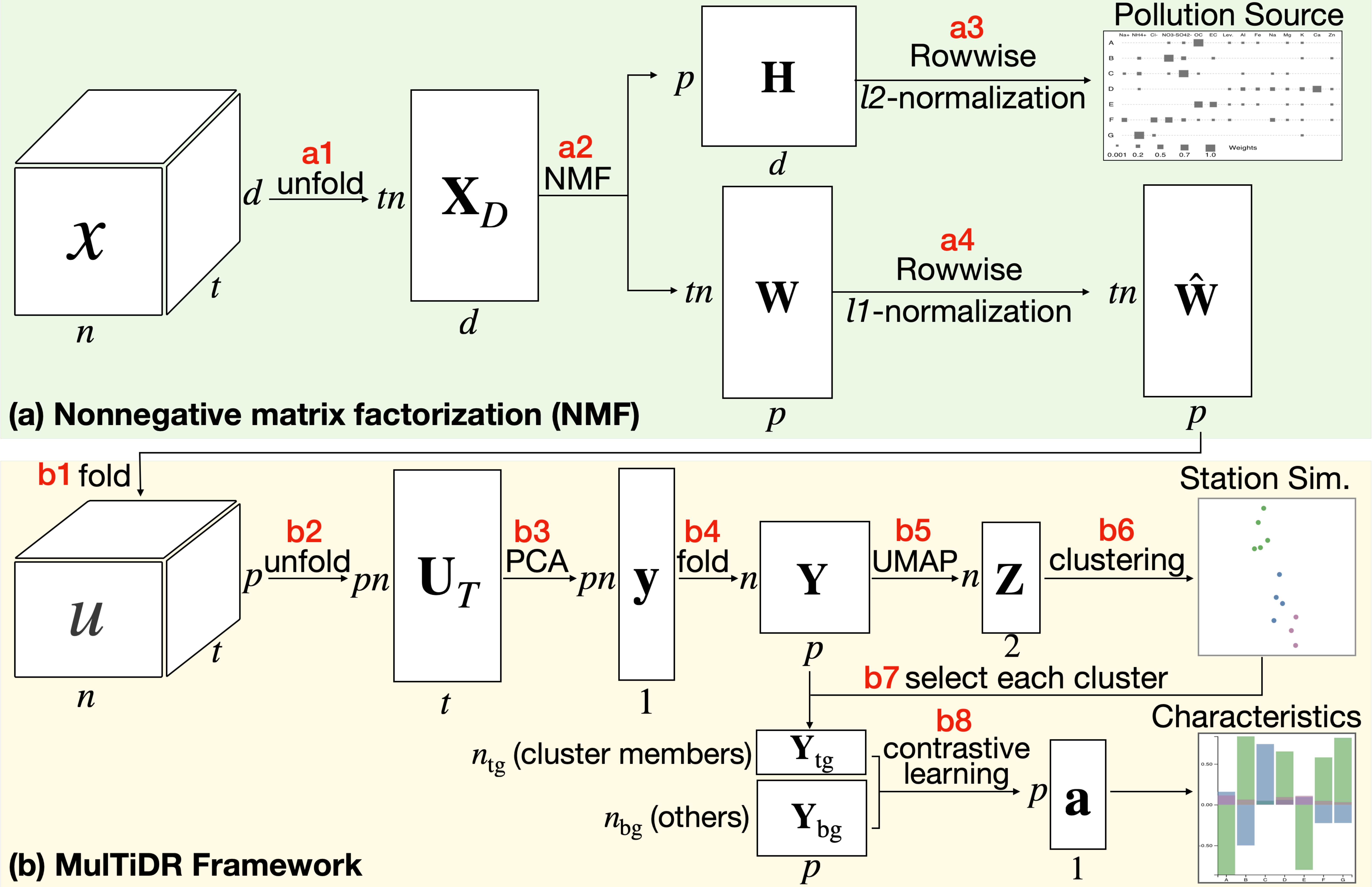}
 \caption{The ML pipeline consisting of (a) NMF and (2) MulTiDR. }
 \label{fig:pipeline}
\end{figure}

\revise{When investigating the relationships among all stations, we want to consider both feature distributions and their temporal changes. 
Unlike approaches that apply DR to only either feature or temporal dimensions, the \MulTiDR{} framework can consider both the feature and time aspects~\cite{fujiwara2020visual}.}
The \MulTiDR{} framework takes a third-order tensor as an input and generates a 2D DR result through the process using two different DR methods, called the two-step DR. 
To use the \MulTiDR{} framework, we first apply tensor folding to  $\NMFLatReprNormal$ and obtain a third-order tensor $\smash{\MulTiDRTensorData \in \mathbb{R}^{\nTimes \times \nInsts \times \nFacts}}$ (\autoref{fig:pipeline}-b1).
In the first DR step, \MulTiDR{} unfolds $\MulTiDRTensorData$ along timestamps and produces a matrix $\smash{\UnfMulTiDRDataT \in \mathbb{R}^{(\nFacts \nInsts) \times \nTimes}}$ (\autoref{fig:pipeline}-b2); then, applies a linear DR method (specifically, we use PCA) to compress the number of dimensions from $\nTimes$ to $1$, resulting in a vector $\Vector{y}$ of $\nInsts \nFacts$ length (\autoref{fig:pipeline}-b3).
In the second DR step, $\Vector{y}$ is first folded to generate a matrix $\Matrix{Y} \in \mathbb{R}^{\nInsts \times \nFacts}$, where each element contains an air pollution source contribution that represents the contributions across all timestamps (\autoref{fig:pipeline}-b4); then, we apply a DR method (specifically, we use UMAP) to generate a 2D plot by reducing the number of dimensions from $\nFacts$ to 2, resulting in a matrix $\Matrix{Z} \in \mathbb{R}^{\nInsts \times 2}$ (\autoref{fig:pipeline}-b5).
$\Matrix{Z}$ shows each station's similarity which considers both feature and temporal aspects. 
We select PCA and UMAP because of their strength in data compression and neighborhood preservation, as discussed in the work of \MulTiDR{}~\cite{fujiwara2020visual}.



To tightly link the feature and space aspects, we perform the first DR on the temporal dimensions of $\MulTiDRTensorData$ and then the second DR on the feature dimensions. 
In this way, the stations' similarities mainly reflect the  distribution of pollution sources' contributions while still considering temporal distribution differences. 
Here, we have briefly introduced the two-step DR in the \MulTiDR{} framework. For the theoretical details, refer to the work by Fujiwara et al.~\cite{fujiwara2020visual}.

After obtaining the 2D DR result, $\Matrix{Z}$, which shows the similarities of stations, we construct groups of stations by applying clustering to $\Matrix{Z}$ (\autoref{fig:pipeline}-b6) and visualize the result in the  {\ViewNameDRScatter} view.
For example, as shown in \autoref{fig:overview}-\ViewLabelDRScatter, 12 stations are clustered in three groups (green, blue, and purple). 
We use $k$-means clustering by default due to its simplicity of parameter selection (i.e., we need to only select the number of cluster $k$) and effectiveness
\suggest{for handling a smaller number of stations.}
However, we can easily change it to the other method (e.g., density-based clustering) based on the data scale and the distribution of point positions in the DR result.


\vspace{-2pt}
\subsubsection{Region Characterization}
\vspace{-1pt}

In addition to the two-step DR, the \MulTiDR{} framework provides the algorithmic support using contrastive learning to interpret the DR result~\cite{fujiwara2020visual,fujiwara2020supporting}. 
Through this interpretation, we can characterize station groups based on their values of features, i.e., pollution sources (\textbf{DG4}).
The \MulTiDR{} framework employs a contrastive learning method called ccPCA~\cite{fujiwara2020supporting}, which reveals features that highly associate to each cluster's uniqueness with respect to others.

The constrastive learning process is shown at the bottom right of \autoref{fig:pipeline}.
Suppose a cluster contains $\nInstsCLTg$ instances, we extract $\smash{\Matrix{Y}_\mathrm{tg} \in \mathbb{R}^{\nInstsCLTg \times \nFacts}}$ by selecting the corresponding rows from $\Matrix{Y}$ (\autoref{fig:pipeline}-b7). 
Then, we also extract the rest of $\Matrix{Y}$ (i.e., other clusters), $\smash{\Matrix{Y}_\mathrm{bg} \in \mathbb{R}^{\nInstsCLBg \times \nFacts}}$, where $\nInstsCLBg = \nInsts - \nInstsCLTg$.
Then, $\Matrix{Y}_\mathrm{tg}$ and $\Matrix{Y}_\mathrm{bg}$ are used as target and background datasets in ccPCA, respectively.
Then, ccPCA produces a weight vector $\smash{\CLFeatCotrib \in \mathbb{R}^{\nFacts}}$, which shows \textit{loadings} of each air pollution source on a cluster's unique characteristics with respect to others (\autoref{fig:pipeline}-b8). 
Similar to loadings in PCA's components, when an absolute loading is large, the corresponding air pollution has high influence on the unique characteristics. 
We repeat this process to obtain $\CLFeatCotrib$ for each cluster. 
Refer to the work by Fujiwara et al.~\cite{fujiwara2020visual} for the details \suggest{on} the contrastive learning process.

As show in \autoref{fig:overview}-\ViewLabelDRCL, we visualize a weight vector $\CLFeatCotrib$ obtained for each cluster in the {\ViewNameDRCL} view, where each bar height, color, and horizontal position represent the loading, cluster ID, and air pollution source, respectively. 
When a bar has a positive loading, a cluster tends to have larger contributions of the corresponding pollution source across different stations and timestamps, and vice versa. 
For example, in the {\ViewNameTimeContrib} view, which we explain in \autoref{sec:time_aspect}, we can see \texttt{Cluster 1} tends to have lower values for \texttt{Source A} than other clusters across timestamps, while \texttt{Cluster 1} tends to have higher values for \texttt{Source B}.
When we have many clusters (e.g., 9 clusters), the overlaps of bar charts could cause a visual clutter.
To alleviate this issue, the system supports filtering based on a cluster ID, which can be applied by hovering over a cluster ID in the cluster legend.

\vspace{-2pt}
\subsubsection{Relating to Geographic Information}
\vspace{-2pt}

We provide the {\ViewNameMap} view (\autoref{fig:overview}-\ViewLabelMap) to inform each station's location along with the clusters obtained with the \MulTiDR{} framework. 
This view can be used to relate the clusters to the geographic locations and understand or validate the clustering result based on the knowledge of the corresponding regions. 
Circular points in \autoref{fig:overview}-{\ViewLabelMap} represent stations' geolocations. 
To overlay the cluster information on stations, we employ Bubble Sets\cite{collins2009bubble}.
The isocontours (or bubbles) are colored by cluster IDs (e.g., the green bubble corresponds to \texttt{Cluster 1}).
Also, the view shows \PM{2.5} values measured at a large number of sensors (specifically, 537 AirBox sensors in our data) to provide general air pollution information (represented by \PM{2.5}) with a high spatial granularity.
We use a rectangle grid, in which we show the average of \PM{2.5} values recorded at stations in each grid at a selected timestamp in the {\ViewNameTrend} view and colorcode the value with a yellow-red colormap.
The same color encoding is also applied to the circular points.


From the visualizations in \autoref{fig:overview}-{\ViewLabelDRScatter}, {\ViewLabelDRCL}, {\ViewLabelMap}, we further understand \texttt{Source F} with the spatial aspect (related to \texttt{Tasks b2}, \texttt{c1}, and \texttt{d1}). 
First, from the {\ViewNameDRCL} view, we notice that \texttt{Source F} has a high contribution to characterizing \texttt{Cluster 1} (colored green). 
Additionally, by looking at the {\ViewNameMap} view, we can see that  \texttt{Cluster 1} is located relatively close to the ocean. 
As a remaining analysis question, we further want to know when \texttt{Source F} dominates the air pollution in the region. 


\vspace{-1pt}
\subsection{Temporal Pattern Extraction}
\vspace{-1pt}
\label{sec:time_aspect}

To relate the air pollution source and region information with the \suggest{temporal} information, we develop two different visualizations: the {\ViewNameTimeContrib} (\autoref{fig:overview}-{\ViewLabelTimeContrib}) and {\ViewNameTrend} (\autoref{fig:overview}-{\ViewLabelTrend}) views.
While the former is to reveal general temporal patterns for each pollution source and region, the latter is to review temporal changes in the main air pollutant, \PM{2.5}, as well as to associate each air pollution source with \PM{2.5}.  

\vspace{3pt}
\noindent\textbf{Analysis of temporal patterns of pollution sources.} 
The {\ViewNameTimeContrib} view (\autoref{fig:overview}-{\ViewLabelTimeContrib}) provides a small multiple of line charts.
Each row of the view corresponds to one pollution source. 
$x$- and $y$-axes represent a timestamp and a contribution of each pollution source, respectively. 
To provide the analysis and visualization scalabilities, instead of showing values of each pollution source for all stations, we visualize an aggregated value for each group of stations. 
To aggregate values, from $\NMFLatReprNormal$, we compute the average contribution of stations in each group at each timestamp.  

From this view, 
around March 18th, \texttt{Source F} has a peak and \texttt{Cluster 1} (green) has an especially high contribution of \texttt{Source F}.  
Thus, we can expect that some pollution events relating to  \texttt{Source F} occurred around this date and heavily affected a region corresponding to \texttt{Cluster 1} (\texttt{Tasks e1} and \texttt{e2}).



\noindent\textbf{Relating the temporal patterns of pollution sources and PM$_\mathbf{2.5}$.} 
The {\ViewNameTrend} view (\autoref{fig:overview}-{\ViewLabelTrend}) uses gray lines to show selected stations' \PM{2.5} values over time.
For example, in \autoref{fig:overview}-{\ViewLabelTrend}, the five stations belonging to \texttt{Cluster 1} are selected. 
Note that there are unconnected lines because the sensors sometimes \suggest{failed} to measure \PM{2.5} due to hardware errors. 
As there are many timestamps for \PM{2.5} (recorded every hour), this view provides a zooming-in/out interaction along the time axis.  
In \autoref{fig:overview}-{\ViewLabelTrend}, we zoom into the days around March 18th (specifically, from 15th to 25th). 
We also show the information of the selected time range at the bottom side of the line chart, where the range from March 15th to 25th is highlighted with a gray rectangle.
Also, the user can select a specific timestamp using a mouse, \suggest{with} the selected timestamp \suggest{shown} at the top left (e.g., March 17th, 10 PM in \autoref{fig:overview}-{\ViewLabelTrend}). 
\PM{2.5} values of all the stations at the selected timestamp are also visualized at the {\ViewNameMap} view.
The gray lines in the {\ViewNameTrend} view can be used to find anomalies or similarities in \PM{2.5} values at some stations and/or some timestamps (\texttt{Task f1}).
For example, we can see that all the stations in \texttt{Cluster 1} have similar \PM{2.5} values in the selected time range.

Additionally, temporal changes of a selected pollution source's contribution to each station are shown with a black line \suggest{(e.g., \texttt{Source F} is selected in \autoref{fig:overview}-{\ViewLabelTrend})}.
While we do not know the direct influence of each pollution source on \PM{2.5} values, this visualization is helpful to see a potential association between each of the sources and \PM{2.5} as those sources consisting of the chemical species often produce \PM{2.5}. 
To further emphasize time periods where the selected pollution source has the highest contribution among all the sources 
for each station, we fill the underneath of a black line within such time periods with a semitransparent black.
Consequently, we see darker-colored areas when more stations share such time periods.

By interactively filtering the stations, within \texttt{Cluster 1}, we observe that \texttt{Source F} has the highest contribution on three stations, \texttt{XC}, \texttt{EL}, and \texttt{DC}, from March 17th to 19th. 
These three stations are located along the coast; thus, this result also supports our interpretation that \texttt{Source F} is sea breezes (\texttt{Tasks f2}, \texttt{f3}).

\vspace{3pt}
\noindent\textbf{Summary of the analysis example.}
We performed one analysis example to explain our system's functionalities and analysis workflow. 
First, based on the NMF result, we identified multiple pollution sources from the data and considered \texttt{Source F} as sea breeze (\texttt{Tasks a1}, \texttt{a2}).
Then, incorporating the aspect of space, we saw \texttt{Source F} has a high influence on the region of  \texttt{Cluster 1}, which contains the stations near the coast (\texttt{b2}, \texttt{c1}, \texttt{d1}).
Lastly, by adding the temporal aspect, we identified when and where in \texttt{Cluster 1} \texttt{Source F} highly contributes to the air pollution---on the three stations along the coast around March 18th (\texttt{Tasks e1}, \texttt{e2}, \texttt{f2}, and \texttt{f3}).
This analysis example demonstrates the importance of analyses from the three aspects and the effectiveness of our flexible workflow, ML pipeline, and the system interface to perform such analyses.

\vspace{3pt}
\noindent\textbf{Implementation.} 
Our system is a web application. For the back-end, we use Python to integrate all the existing algorithms of NMF and \MulTiDR{}. Specifically, we use Scikit-Learn\cite{pedregosa2011scikit} implementation for NMF, where we adopt the nonnegative double singular value decomposition for the consideration of data sparsity. 
For \MulTiDR{}, we use the implementation the original authors provided~\cite{fujiwara2020visual}. 
To implement the front-end user interface, we use a combination of HTML5, JavaScript, React, and D3\cite{bostock2011d3}. We use Django and GraphQL to communicate between the front-end and back-end.
\vspace{-1pt}
\section{Use Cases}
\vspace{-1pt}
\label{sec:cases}



\autoref{sec:methodology} has already demonstrated a concrete case using one subworkflow. 
Here, using other subworkflows, we present three additional analyses on the air pollution data obtained in Taiwan (Cases 1, 2) and another open dataset obtained in the United States (Cases 3, 4). 

While these cases are performed by our team's visualization researchers who have gained basic domain knowledge through literature study on the air pollution analysis research, the findings in each analysis (including the one in \autoref{sec:methodology}) are reviewed and evaluated by the domain researchers in our team.
For example, they expressed that our findings on \texttt{Source F} in \autoref{sec:methodology} are very interesting. 
From the \ViewNameNMFResult{} view, they also found that \texttt{Source F} shows a high concentration of $\mathrm{NO}_{3}^{-}$ (i.e., nitrate) as well as \ce{Na+} and \ce{Cl-}.
This composition was likely caused by an atmospheric reaction between \ce{NaCl} and \ce{HNO3}, indicating that \texttt{Source F} represents aged sea salt with nitrate contamination~\cite{chou2008implications}.
They also showed positive impressions on our visual analytics system's functionalities and are interested in applying our system to their ongoing research projects.


\vspace{-2pt}
\subsection{Case\,1:\,Explaining\,the\,Cause\,of\,Air\,Pollution\,Events}
\vspace{-1pt}
\label{sec:case-explain}

\begin{figure}[tb]
	\centering
    \includegraphics[width=\linewidth]{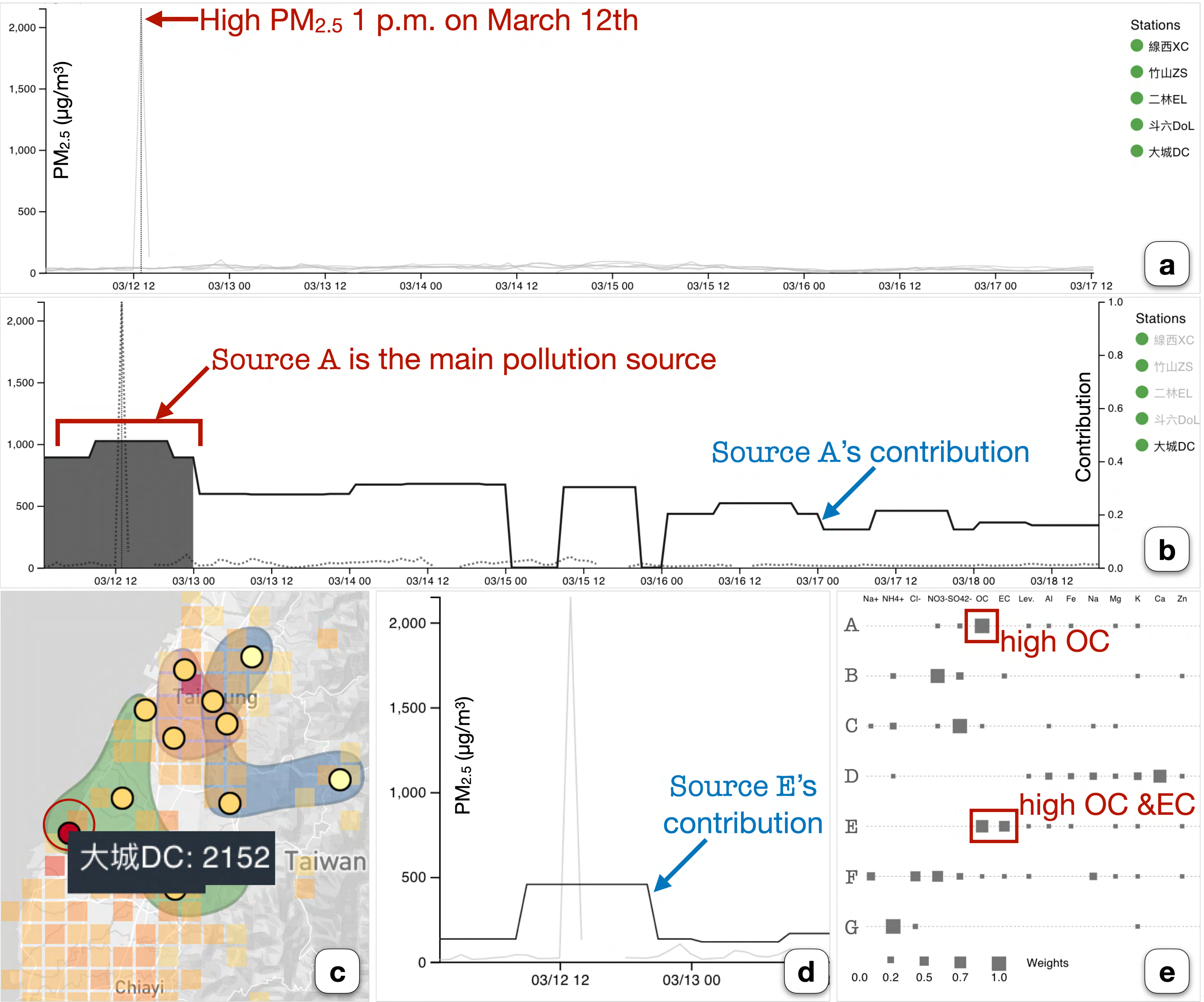}
    \caption{Case 1. Explaining the cause of the air pollution event: (a) the anomaly inspection of PM$_{2.5}$, (b) the contribution of \texttt{Source A} to the anomaly in \texttt{Station DC}, (c) the location of the corresponding anomaly, (d) the contribution of \texttt{Source E} to the anomaly in \texttt{Station DC}, and (e) the source interpretation with the chemical species.}
	\label{fig:case-anomaly}
\end{figure}


This subworkflow starts by anomaly inspection, a common task for domain experts in analyzing air pollution data, followed by understanding the cause of the anomaly through identification of the contributing pollution source to the anomaly. 

As annotated in \autoref{fig:case-anomaly}-a, in the \ViewNameTrend{} view, we observe a \PM{2.5} anomaly with a value of \SI{2,152}{\micro\gram/\cubic\metre} at 1:00 p.m. on March 12th, 2018 (\texttt{Task f1}). 
To find a station that recorded this anomaly, we hover over the \ViewNameTrend{} view with a vertical auxiliary line indicating the selected timestamp. 
Then, as shown in \autoref{fig:case-anomaly}-c, we inspect the \ViewNameMap{} view, which visualizes the observed \PM{2.5} values at the selected timestamp, and we visually identify that the anomaly is recorded by \texttt{Station DC} (\texttt{Task d2}).
We focus on \texttt{Station DC} in the \ViewNameTrend{} view and manually select each of the pollution sources to review their  contributions to \texttt{Station DC} (\texttt{Task f2}).
When selecting \texttt{Source A} (\autoref{fig:case-anomaly}-b), we observe that \texttt{Source A} has the highest contribution during the corresponding time period, whereas \texttt{Source E} (\autoref{fig:case-anomaly}-d) has the second highest contribution (\texttt{Task f3}). 
This indicates that \texttt{Sources A} and \texttt{E} are likely related to this anomaly event.

As shown in \autoref{fig:case-anomaly}-e, using the \ViewNameNMFResult{} view, we further examine chemical species that have high concentrations in \texttt{Sources A} and \texttt{E} (\texttt{Task a2}). 
As annotated in \autoref{fig:case-anomaly}-e, we can see that organic carbon (abbreviated as OC) has the highest concentration in \texttt{Source A}, while \texttt{Source E} has high concentrations of both OC and elemental carbon (EC). 
Because EC comes from the incomplete combustion of carbon-contained materials and OC can be produced through the organic compound gas-to-particle process, \texttt{Sources A} and \texttt{E} are likely related to biomass burning.
\textcolor{blue}{The domain researchers confirmed that this case was due to biomass burning. 
They explained that the existence of Levoglucosan in \texttt{Sources A} and \texttt{E} (even with a small concentration) further supports this finding as Levoglucosan is a species that is almost totally derived from biomass burning~\cite{simoneit1999levoglucosan}.}

Through this subworkflow (\texttt{Tasks f1}, \texttt{d2}, \texttt{f2}, \texttt{f3}, and \texttt{a2}), we may infer that biomass burning caused this air pollution event at \texttt{Station DC}. 
According to the Environmental Protection Agency (EPA) and the local fire department~\cite{bioburnDC}, the related air pollution was caused by the household straw combustion at this time period. 
Further, Chen et al.~\cite{chen2017review} discussed different types of biomass burning and indicated that OC and EC often associate with straw combustion emissions. 
This information validates not only the factorization result by NMF but also the inference derived from this subworkflow.


\vspace{-2pt}
\subsection{Case 2: Uncovering Hidden Air Pollution Events}
\vspace{-1pt}

\label{sec:case-hidden}


\begin{figure}[tb]
	\centering
    \includegraphics[width=\linewidth]{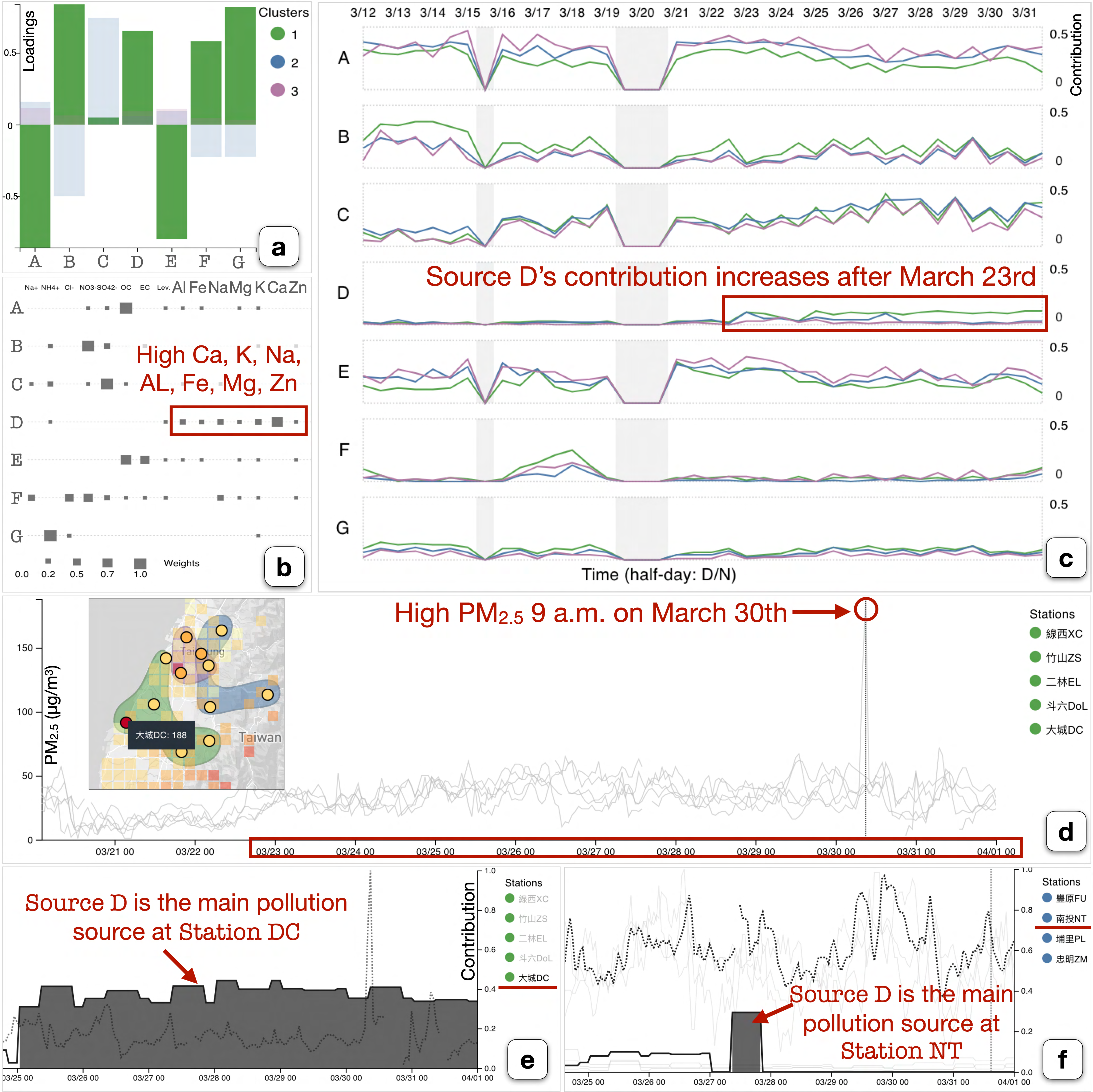}
    \caption{Case 2. Uncovering hidden air pollution events: (a) the inspection of influential sources on each station group's characteristics, (b) the examination of time periods with high pollution source contributions, (c) the observation of PM$_{2.5}$ values during the period for \texttt{Cluster 1}, (d) the contribution of \texttt{Source D} to \texttt{Station DC}, (e) the interpretation of \texttt{Source D} with the chemical species, and (f) the contribution of \texttt{Source D} to \texttt{Station NT} in the same period.}
	\label{fig:case-hidden-event}
\end{figure}


This subworkflow starts by examining the characteristics of one region and by finding a pollution source of interest.
Then, the subworklow moves on to a detailed exploration of a time period where the source highly contributes to the pollution in order to uncover any hidden air pollution events in the transition. 

Here we focus on \texttt{Cluster 1}, which includes \texttt{Station DC} we have examined in Case 1.
From the \ViewNameDRCL{} view shown in \autoref{fig:case-hidden-event}-a, we notice that many sources have high absolute loadings on the characteristics of \texttt{Cluster 1} (\texttt{Task c1}). 
We check each pollution source's contribution transitions in \autoref{fig:case-hidden-event}-c. 
From this view, we can see that each source has different contribution peaks and \texttt{Cluster 1} has different levels of peaks from the other clusters for all the sources except for \texttt{Source C} (\texttt{Task e3}), which is expected from \autoref{fig:case-hidden-event}-a showing \texttt{Source C}'s minor loading on \texttt{Cluster 1}.
One interesting observation is that, as annotated by the red rectangle, \texttt{Source D} starts to contribute to the air pollution after March 23rd and it keeps having a much higher contribution to \texttt{Cluster 1} when compared it to the other clusters (\texttt{Tasks e1}, \texttt{e2}).
This sudden increase in the contribution implies the possible existence of pollution events around that period, which leads us to use the \ViewNameTrend{} view for a more detailed examination.

As shown in \autoref{fig:case-hidden-event}-d, we visualize \PM{2.5} values of the stations in \texttt{Cluster 1}.
We see only the slight increase of \PM{2.5} at each station after around March 23rd except for one acute peak at 9:00 a.m. on March 30th, as annotated with the red circle (\texttt{Task f1}).
However, since our previous finding from the \ViewNameTimeContrib{} view suggests that \texttt{Source D} increases its contribution around this period, we further inspect the contribution of the \texttt{Source D} to the stations in \texttt{Cluster 1} with the \ViewNameTrend{} view.
As identified by the filled areas in \autoref{fig:case-hidden-event}-e, after interactively filtering the stations, we can see that \texttt{Source D} has the highest contribution to \texttt{Station DC} from March 25th to March 31st (\texttt{Task f3}). 

According to the local fire department's report on March 30th, 2018~\cite{jossburnDC}, there were 180 fire incidents around \texttt{Station DC} from March 24th to 30th. 
They indicated that these were all due to the forthcoming Qingming festival. 
Since the burning  of some papers, called joss papers, as a ritual offering is an important part of the festival, fire incidents frequently happen, especially in rural areas. 
The \PM{2.5} anomaly observed at \texttt{Station DC} at 9:00 a.m. on March 30th also has been reported and investigated by the EPA~\cite{jossburnDC}, where the cause is confirmed to be the joss paper burning. 

Khezri et al.~\cite{khezri2015annual} studied the impact of the joss paper burning on air pollution. 
Their results show that the increase in the concentrations of \ce{Zn}, \ce{Ca}, \ce{K}, \ce{Mg}, \ce{Fe}, and \ce{Al} (listed in order of the increase) can be linked to the increase of \PM{2.5} during the joss paper burning. 
They further reported that \ce{Ca}, \ce{Al}, \ce{Na}, \ce{Mg}, and \ce{K} (in order of the concentration) have a high concentration in the unburnt joss paper.
As shown in the \ViewNameNMFResult{} view (\autoref{fig:case-hidden-event}-b), we observe that \texttt{Source D} is mostly characterized by \ce{Ca}, \ce{K}, \ce{Na}, \ce{Al}, \ce{Fe}, \ce{Mg}, and \ce{Zn} and is thus likely related to the combustion of joss paper (\texttt{Task a2}). 
Without the information obtained from the \ViewNameDRCL{} and \ViewNameTimeContrib{} views, we would not have be able to reach this finding.
This shows the usefulness of our ML pipeline to find this type of hidden pollution events.

We can further inspect the contribution of \texttt{Source D} to the other clusters in order to investigate any possible hidden pollution events (\texttt{Tasks f2}, \texttt{f3}). 
For example, as shown in \autoref{fig:case-hidden-event}-f, we observe that \texttt{Source D} has the highest impact on \texttt{Station NT} belonging to \texttt{Cluster 2} on March 27th, indicating another possible pollution event due to the joss paper burning for the Qingming festival.

In this analysis involving many tasks in our flexible workflow, we can infer that \texttt{Source D} characterizes \texttt{Cluster 1}, as it highly contributes to  \texttt{Station DC} from March 25th to 31st, which is not visible if observing only \PM{2.5} values. 
More specifially, \texttt{Source D} is likely related to the joss paper burning. 
With this potential relation in mind, we may explore other regions to examine the possible existence of hidden air pollution events, such as one potential event at \texttt{Station NT} on March 27th.


\vspace{-2pt}
\subsection{Case 3: Validating Data Quality}
\vspace{-1pt}

\label{sec:case-us}

\begin{figure}[tb]
	\centering
    \includegraphics[width=\linewidth]{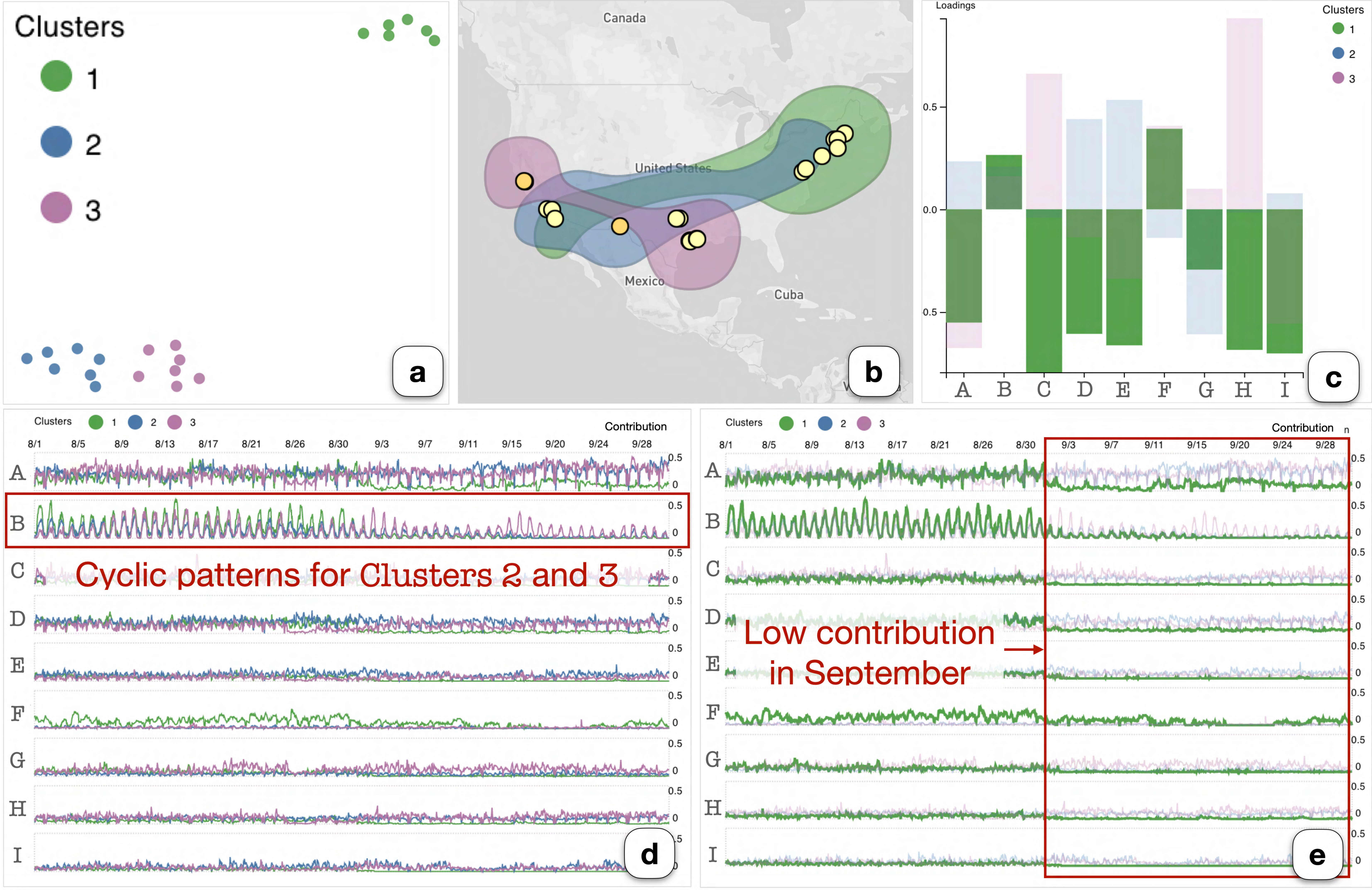}
    \caption{Case 3. Validating data quality: (a) the grouping of similar stations, (b) the comparison of the clustering result with the geographic information, (c) the inspection of influential sources on \texttt{Cluster 1}'s characteristics, (d) the examination of source contribution transition, (e) the observation of a clear change in the contributions to \texttt{Cluster\,1}.
    }
	\label{fig:case-us}
\end{figure}


This subworkflow starts with the validation of the clustering result using the \ViewNameDRScatter{} view and the \ViewNameMap{} view.
While the clustering using the \MulTiDR{} framework incorporates all of the three aspects, the geographical locations provide us a good intuition to assess clustering quality as adjacent stations usually have similar influence from air pollutants.
Thus, by comparing these two views, we can perform quick validation of data quality from multiple aspects (utilizing the \MulTiDR{} framework) in an intuitive manner (referring to the geographical locations).
When observing suspicious clustering results, we can further identify the cause of the issue in a top-down manner, such as finding influential sources on clusters and then reviewing the space or time aspect of such sources.



In this use case, we analyze the air pollution open dataset provided by the U.S. EPA~\cite{usopendata} to demonstrate our system's applicability to various datasets. 
We use 20 stations that collect both \PM{2.5} and 57 VOCs (e.g., benzene, ethane, and isoprene). 
All of these values are recorded on an hourly basis, from August 1st to September 30th, 2020. 
This data forms a third-order tensor $\smash{\TensorData \in \mathbb{R}^{\nTimes \times \nInsts \times \nFeats}}$, where $\nTimes \!\!=\!\! 1,464$, $\nInsts \!\!=\!\! 20$, and $\nFeats \!\!=\!\! 57$. 
For NMF, we select $p=9$.


After loading the dataset into the system, we first review the relationships among stations in the \ViewNameDRScatter{} view (\autoref{fig:case-us}-a).
This view shows that \texttt{Cluster 1} is well separated from the others (\texttt{Tasks b1} and \texttt{b2}). 
Thus, we can expect that \texttt{Cluster 1} has different behaviors from the others (in terms of the feature and time aspects as described in \autoref{sec:region}). 
However, as shown in \autoref{fig:case-us}-b, many stations in \texttt{Cluster 1} are located with similar regions to \texttt{Cluster 2}. 
This is counterintuitive as stations at different locations are clustered together (\texttt{Task d1}). 

To investigate the cause of this cluster formation, we inspect the \ViewNameDRCL{} view (\autoref{fig:case-us}-c), and we notice that most sources have negative loadings on the characteristics of \texttt{Cluster 1} (\texttt{Task c1}). 
We then check each pollution source's contribution transition in \autoref{fig:case-us}-d and highlight \texttt{Cluster 1} (\autoref{fig:case-us}-e). 
We observe that, starting from September, most sources keep having a much lower contribution to \texttt{Cluster 1} when compared with the other clusters in the same period and even \texttt{Cluster 1} before September (\texttt{Tasks e1,} \texttt{e2}, and \texttt{e3}). 
As some sources (e.g., \texttt{Source I}) even show no contribution to \texttt{Cluster 1} in September, we check the raw data outside the system and then find out that, during this period, the stations in \texttt{Cluster 1} have many missing values for the VOCs related to those sources. 
We thus infer that the missing values in September lead to this cluster formation. 
This result notifies us that, during analyses on this data, we should keep being aware of potential analysis issues due to these missing values.

\begin{figure}[tb]
	\centering
    \includegraphics[width=\linewidth]{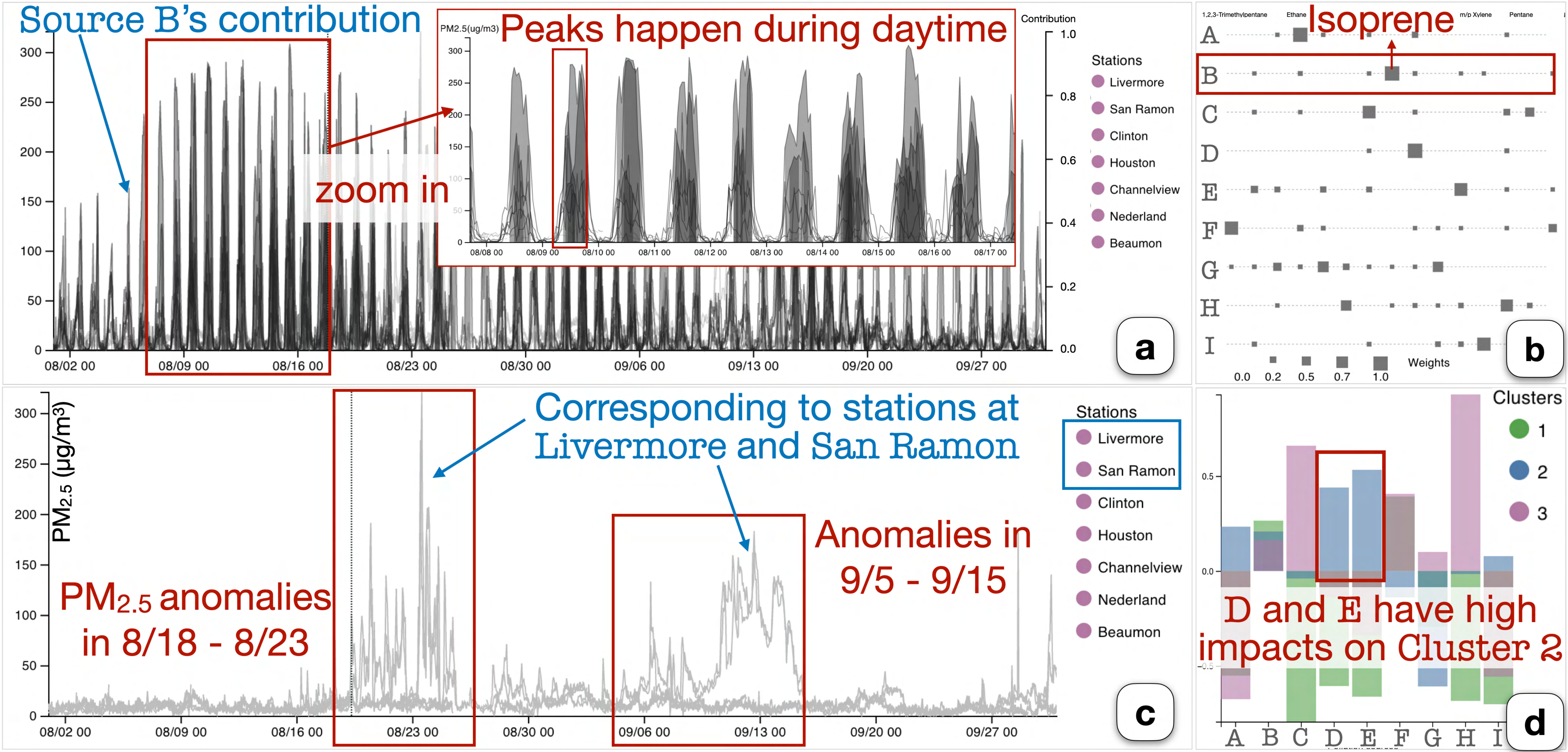}
    \caption{Case 4. Combining multiple subworkflows: (a) the investigation of the cyclic contribution of \texttt{Source B} to \texttt{Cluster 3} after the data validation in \autoref{fig:case-us}, (b) the interpretation of \texttt{Source B} with the chemical species, (c) the observation of PM$_{2.5}$ values of the stations in \texttt{Cluster 3}, and (d) the inspection of \texttt{Sources D} and \texttt{E} as influential sources on \texttt{Cluster 2}'s characteristics.}
	\label{fig:case-us2}
\end{figure}

\vspace{-1pt}
\subsection{Case 4: Combining Multiple Subworkflows}
\vspace{-1pt}
\label{sec:case-us2}

In the previous cases, we have separately demonstrated different subworkflows.
However, as the analysis workflow supported by the system is highly flexible, we can connect multiple subworkflows (resulting in a more all-inclusive subworkflow) to thoroughly analyze the data. 
Here, following the data validation performed in Case 3, we analyze the U.S. air quality data mainly from the time aspect.

As we have noticed the data quality issue in \texttt{Cluster 1}, we focus on analyzing \texttt{Clusters 2} and \texttt{3}.
Going back to \autoref{fig:case-us}-d, we see that \texttt{Source B} has a continuous cyclic pattern for both \texttt{Clusters 2} and \texttt{3} (\texttt{Task e1}). 
We further assess this pattern with the \ViewNameTrend{} view.
As shown in \autoref{fig:case-us2}-a, we observe that \texttt{Source B} usually has a high contribution during the daytime (\texttt{Tasks f2}). 
We then examine the \ViewNameNMFResult{} view to interpret which chemical species are related to \texttt{Source B} (\autoref{fig:case-us2}-b). 
We see only isoprene has a high concentration in \texttt{B} (\texttt{Task a2}). 
Following the study by Brown et al.~\cite{brown2007source} and the guidelines by the U.S. EPA~\cite{epabiogenic}, we identify that \texttt{Source B} is likely biogenic emissions, a natural pollution source often contributed by photochemical reactions of vegetation. 

We also review \PM{2.5} values of the stations in \texttt{Clusters 2} and \texttt{3} with the \ViewNameTrend{} view.
As annotated in \autoref{fig:case-us2}-c, when selecting \texttt{Cluster 3}, we observe that there are \PM{2.5} anomalies at \texttt{Stations Livermore} and \texttt{San Ramon} at two different time periods. 
We can thus use the same subworkflow described in Case 1 to understand the causes. 
Similarly, from \autoref{fig:case-us2}-d, we can explore how \texttt{Sources D} and \texttt{E} influence \texttt{Cluster 2} using the same subworkflow described in Case 2. 
This case shows how we can transition from one subworkflow to another based on intermediate analysis findings, demonstrating the flexibility of our analysis workflow. 

\vspace{-1pt}
\section{\revise{Expert Review}}
\vspace{-1pt}
\label{sec:expertreview}

\revise{To evaluate the usefulness of our prototype system, we conducted an informal interview with two experts in air pollution analysis. 
The first expert (E1) is an assistant professor in a public health department who has knowledge of the dataset used in \autoref{sec:case-explain} and \autoref{sec:case-hidden}.
The second expert (E2) is a researcher with over 10 years of experience in air quality simulation and modeling.
To facilitate discussion from various perspectives, the interview as conducted jointly by two of our team members: one is an air quality scientist and the other is a visualization expert.
The informal interview used a video conference setup. 
We first explained our work's background (including the design goals) and our system; then, we demonstrated the two use cases described in \autoref{sec:case-explain} and \autoref{sec:case-hidden}. 
While our team's visualization researcher operated the system remotely, all the participants were able to specify what user actions they would like to perform during the interview.}

\revise{
All the participants confirmed that the design goals fulfill the analysis requirements and the system achieved all the set goals. 
Generally, they had positive comments on how the system enables efficient and effective analyses. 
E1 commented, ``\textit{This system allows us to simultaneously examine how pollution source contribution changes over time and across stations, which was not available before.}''
They also stated that the cluster information in the \ViewNameMap{} view is useful, as E2 noted, ``\textit{All the air pollution analysis eventually requires our geographical understanding for validation.}''}

\revise{
The two experts also pointed out potential improvements on the system. E1 said, ``\textit{The use cases seem reasonable. I can easily tell Sources A and E are related to biomass burning, but it is hard to identify their differences. There might be tracers that exclusively exist in one of them}'', 
and
suggested that statistical support could be provided to inform the reliability of the NMF results.
E2 asked, ``\textit{How would the stations' similarities change if we also consider the wind information?}'' and was further interested in the wind influence on the pollutant propagation.
} 

\revise{
Overall, all the participants agreed that the system provides essential information for performing analysis and each visual component is easy to use.
They noted that their familiarity with the studied environment might lead to this impression; however, they also commented that, in any case, such familiarity is required to perform air pollution analysis.
We plan to incorporate their suggestions in our future work, for example, by making the system more flexible to perform analyses related to the wind field. 
}

\vspace{-2pt}
\section{Discussion}
\vspace{-2pt}

Through the analysis examples and the expert review, we have shown the effectiveness of our system and flexible workflow for data-driven air pollution analysis.
Here, we provide additional discussions on our visual interface, ML pipeline, and flexible analysis workflow.  

\vspace{3pt}
\noindent
\textbf{Visual scalability.}
As discussed in \autoref{sec:methodology}, our system provides better analysis scalability for all of the three aspects by incorporating multiple ML methods. 
For example, the number of features is reduced by NMF and the \MulTiDR{} framework clusters stations while considering both time and feature aspects. 
With the aid of ML, we can effectively and efficiently analyze air pollution data with the scale of those currently studied by domain researchers (e.g., the data analyzed in \autoref{sec:cases}).
However, we can expect that, in the future, we would be able to collect larger scale data in terms of data variety (with the advancement of the sensing technology) and volume (with the reduction of the device cost). 
As have been seen in the ML field, ML methods would still deal with such higher variety and volume data, but visual scalability could limit the analysis capability.

For example, for the feature aspect, our system currently supposes the use of a small number of pollution sources (e.g., less than 10); however, in the future, the number of available chemical species would be increased and, as a result, a larger number of pollution sources would need to be extracted (e.g., 30). 
In such a case, the \ViewNameNMFResult{} view should use a more scalable visualization, such as a heatmap that aggregates rows based on their similarities~\cite{fujiwara2020supporting}. 
For the space aspect, as the number of clusters increases, the \ViewNameMap{} view suffers from visual clutters due to the Bubble Sets conveying the cluster information.
To solve this, alternative visualization designs should be developed in the future. 
For the time aspect, when the temporal granularity is increased (e.g., recording every tenth minute), directly visualizing time series would not be helpful to find trends because of noises and fluctuations in the recorded data. 
For this, we can consider to further incorporate computational analysis methods developed for temporal data, such as smoothing and functional data analysis methods~\cite{mvod}.  

\vspace{3pt}
\noindent
\textbf{Consideration of information loss and spatiotemporal continuity.}
Our ML pipeline sequentially uses three different DR methods (NMF, PCA, and UMAP) to extract the essential information. 
The information loss could be inherited and amplified through each DR. 
However, as demonstrated in case studies, our methodology is still able to provide insights since each of the three DR methods reduces unimportant information with a different focus. 
NMF considers the co-occurrences of chemical species to identify a pollution source, PCA tries to preserve the variance of the temporal distribution for each pollution source, and UMAP retains the similarities of the source behaviors for stations. 
Also, the amount of information loss can be captured by measures provided by these DR methods, such as ``explained variance ratio'' in NMF and PCA.
When the loss is too high \revise{(e.g., in a case with many timestamps)}, we can consider applying feature selection before DR.

One limitation of the current ML pipeline is the lack of consideration of the continuity among different time points and locations. 
For example, air pollution usually propagates from one place to another.
\suggest{Our system employs NMF because it is more familiar with domain experts than advanced decomposition methods, which consider the continuity of space, time, or both.}
However, to provide further analysis capabilities, we plan to provide an option to use advanced methods, such as dynamic mode decomposition~\cite{schmid2010dynamic}.

\vspace{3pt}
\noindent
\textbf{Guiding analyses.} 
Our methodology implements the flexible analysis workflow by coupling the ML pipeline and linked visualizations. 
As demonstrated, with this methodology, we can complete various analysis tasks to derive insights from the data. 
Our interface even supports the flexible coordination of views, allowing domain experts to focus on a specific set of visualizations based on their ever-changing analysis needs.
Unlike visual analytics systems following Shneiderman's mantra---``overview first, zoom and filter, then details-on-demand''\!~\cite{shneiderman1996eyes}, our workflow does not have one-directional analysis steps\suggest{. Instead, m}ultiple views provide overviews or details but with \suggest{a} different focus (e.g.,  the \ViewNameNMFResult{} and \ViewNameTimeContrib{} views provide the overviews of features and time points, respectively).
\revise{The analysis is guided by visually distinct elements (e.g., a high bar in the \ViewNameDRCL{} view suggests an important pollution source to understand clusters), which is close to the concept of ``entry point'' in the HCI field~\cite{kirsh2001context}.}

Before being familiar with analysis using our system, domain experts might not be able to decide which view to begin or move on to, in order to successfully perform their analysis tasks. 
\suggest{W}e plan to extract common analysis tasks and the corresponding subworkflows through long-term use of our system by domain experts; then, analogous to map applications, based on the selected analysis task (i.e., destination), visually guide analysis steps (i.e., path) to complete the task.
Also, to uncover patterns in multivariate, spatiotemporal data, broad exploration of the data might be needed.
By extending existing approaches~\cite{xu2018chart}, we plan to inform which part of the data is already or still not investigated.
\revise{We can further investigate how these guides can alleviate analysts' cognitive loads.}

\vspace{-2pt}
\section{Conclusion}
\vspace{-2pt}

We have introduced a visual analytic system that supports a variety of analysis tasks for air pollution data analysis. 
By incorporating multiple machine learning methods, the system enables the efficient exploration of the data and provides a flexible analysis workflow that involves domain experts in the loop.
As demonstrated in the use cases, our system's analysis capability aids in performing various domain-specific analyses.
Our work, thus, contributes to facilitating data-driven air pollution analysis. 

\acknowledgments{
This research was supported in part by the U.S. National Science Foundation through grant IIS-1741536 and a gift grant from Bosch Research. 
}

\bibliographystyle{abbrv-doi}

\bibliography{main}

\end{document}